\documentclass[11pt]{article}
\usepackage{amsmath,amssymb,graphicx}
\usepackage{booktabs,makecell,multirow,tabularx}
\usepackage{dblfloatfix}
\usepackage{microtype}
\usepackage{xurl}
\usepackage[preprint]{acl}
\usepackage{times}
\usepackage{latexsym}
\usepackage[T1]{fontenc}
\usepackage[utf8]{inputenc}
\usepackage{inconsolata}

\author{
  Akshit Sharma \and Prashant W. Patil \\
  CVPR Lab, MFSDSAI \\
  Indian Institute of Technology Guwahati \\
  \texttt{akshitsharma.rs@gmail.com} \quad
  \texttt{pwpatil@iitg.ac.in}
}
\title{What Improves Multimodal Misinformation Detection? Answers from a Large-Scale Empirical Study}

\begin{document}
\setlength{\emergencystretch}{2em}
\setlength{\textfloatsep}{8pt plus 1pt minus 2pt}
\setlength{\dbltextfloatsep}{8pt plus 1pt minus 2pt}
\setlength{\floatsep}{6pt plus 1pt minus 2pt}
\setlength{\intextsep}{6pt plus 1pt minus 2pt}
\setlength{\abovecaptionskip}{4pt}
\setlength{\belowcaptionskip}{0pt}
\maketitle

\begin{abstract}
Multimodal misinformation is increasingly crafted to look convincing by pairing a textual claim with an image that appears to ``prove'' it. Yet in practice, building effective detectors often hinges on a small set of design choices that are rarely examined in a controlled way. In this paper, we conduct a large-scale study of multimodal design choices for misinformation detection with over 3,375 experiments- spanning three benchmark datasets and a broad range of pre-trained vision and language backbones. Through systematic comparisons and targeted robustness analyses, we distill practical guidance on which design choices help, when do they fail silently, and what aspects of the pipeline most strongly shape model behavior, answering 4 key Research Questions (RQs). We aim to provide a reliable foundation for designing stronger and more dependable multimodal misinformation detection systems, thus contributing to the broader research community.
\end{abstract}

\section{Introduction} \label{sec:intro} Misinformation on social media and online news is increasingly multimodal: the core claim is expressed through text, while an accompanying image is often interpreted as corroborating evidence. In practice, however, the image may be irrelevant, out-of-context, weakly related, or intentionally manipulated, yet still persuasive enough to mislead readers. This makes credibility assessment difficult for unimodal systems, since text-only models cannot verify visual evidence, while image-only models cannot interpret what is being claimed. As a result, multimodal misinformation classification has become a standard framing for fake-news detection, rumor detection, and image--text inconsistency analysis~\cite{abdali2024multimodal,comito2023survey,jin2017attrnn,luo2021newsclippings}. A large body of work addresses this setting, but the design space is now broader than fusion alone. Att-RNN~\cite{jin2017attrnn} integrates textual, visual, and social-context signals; EANN~\cite{wang2018eann} targets event-specific bias; SpotFake~\cite{singhal2019spotfake} and SpotFake+~\cite{singhal2020spotfakeplus} show the strength of transfer-learning pipelines built from pre-trained encoders; MVAE~\cite{khattar2019mvae} studies shared latent spaces; and models such as SAFE~\cite{zhou2020safe}, MCNN~\cite{xue2021mcnn}, and recent out-of-context image--text verification work such as NewsCLIPpings~\cite{luo2021newsclippings} and SNIFFER~\cite{qi2024sniffer} explicitly target cross-modal mismatch. Adjacent image--text classification tasks expose similar design tensions. In multimodal meme classification, CLIP-based methods investigate explicit cross-modal interaction and lightweight adaptation of pretrained representations~\cite{kumar2022hateclipper,shah2024memeclip}, while more recent work explores keyword-guided representation learning~\cite{sharma2026memetag} and local--global pivotal-cue grounding with frozen CLIP features~\cite{sharma2026smallcues}. Although these tasks target harm, hate, or sarcasm rather than veracity, they motivate the same broader question of how encoder choice, fusion, and modality reliance shape robust image--text classification.

Across this literature, however, several recurring design factors are typically entangled: the visual backbone, the text backbone, the level of multimodal interaction, the downstream head, and the model's dependence on each modality. Yet a basic question remains unresolved: \emph{when it comes to multimodal misinformation classification, which of these design factors actually contribute to model performance and which don't?} Prior work often varies several factors at once, or reports only the best benchmark-specific configuration, making it difficult to isolate what actually drives performance~\cite{abdali2024multimodal,comito2023survey,chen2023robustness}. This matters because many practical pipelines rely on frozen pre-trained encoders and lightweight heads, where small design decisions can dominate the final result. In this paper, we conduct a controlled study of multimodal misinformation detection that isolates each design component and evaluate the resulting benefits, trade-offs at scale. Concretely, we define three fusion strategies based on the abstraction level being combined---\emph{early} fusion (feature-level), \emph{mid} fusion (score-level), and \emph{late} fusion (decision-level)---and evaluate them as part of the design factors under a unified protocol. We then address four key research questions: \begin{itemize} \item \textbf{RQ1. Generalized Performance Trends:} Which end-to-end performance patterns are seen to remain stable across different large-scale multimodal misinformation benchmarks? \item \textbf{RQ2. Modality Complementarity:} When do text and image contribute complementary evidence, and when does one modality dominate when it comes to multimodal misinformation detection? \item \textbf{RQ3. Impact of Each Component on Overall Performance:} Which pipeline components-visual encoder, text encoder, fusion strategy, or classifier, exert the most reliable effect on the classification performance? \item \textbf{RQ4. Benchmark-Shift Generalization:} To what extent do strong in-benchmark design choices and observed performance trends generalize under transfer to unseen misinformation benchmarks? \end{itemize} Across the three benchmark datasets, we conduct a total of 3,375 experiments- consisting of 675 in-domain model evaluations, 1,350 modality-drop evaluations, and 1,350 cross-dataset transfer evaluations. \noindent\textbf{Contributions.} \begin{enumerate} \item \textbf{RQ-based Analysis:} We provide a unified empirical analysis of performance trends, modality dependence, component robustness, and transfer under benchmark shift, providing answers to the 4 key research questions. \item \textbf{Evidence at scale:} 675 + 1,350 + 1,350 = 3,375 controlled experiments across three benchmark datasets a unified protocol and diverse backbone combinations, supporting reproducible and broadly applicable conclusions. \item \textbf{Statistically grounded conclusions:} We report confidence intervals, paired hypothesis tests, and transfer analyses that separate benchmark-stable findings from dataset-contingent effects. \end{enumerate}
\section{Methodology}
\subsection{Problem Formulation}
Let $\mathcal{D}=\{(x_i^{\text{img}},x_i^{\text{text}},y_i)\}_{i=1}^{N}$ be a multimodal misinformation dataset, where $x_i^{\text{img}}$ is an image, $x_i^{\text{text}}$ is the associated text, and $y_i\in\{1,\dots,C\}$ is the class label. We learn a classifier
\[
f:\mathcal{X}_{\text{img}}\times \mathcal{X}_{\text{text}}\rightarrow \{1,\dots,C\}
\]
using a modular pipeline with a vision encoder $\phi_v:\mathcal{X}_{\text{img}}\rightarrow\mathbb{R}^{d_v}$ and a text encoder $\phi_t:\mathcal{X}_{\text{text}}\rightarrow\mathbb{R}^{d_t}$:
\[
\begin{aligned}
z^{\text{img}} &= \phi_v(x^{\text{img}})\in\mathbb{R}^{d_v},\\
z^{\text{text}} &= \phi_t(x^{\text{text}})\in\mathbb{R}^{d_t}.
\end{aligned}
\]
Fusion combines $(z^{\text{img}},z^{\text{text}})$ into a joint representation, which is then passed to the final classifier.
\subsection{Encoders and Classifiers Used}

\subsubsection{Vision Encoders}
We instantiate $\phi_v$ using five pre-trained vision backbones that span both convolutional and transformer-based families: ResNet-50~\cite{he2016resnet}, ViT-B/16~\cite{dosovitskiy2021vit}, CLIP ViT-B/32~\cite{radford2021clip}, SigLIP~\cite{zhai2023siglip}, and VGG-16~\cite{simonyan2015vgg}. This set covers standard CNNs, vanilla vision transformers, and image encoders trained with image--text pretraining objectives.

This selection is included to make the visual encoder a controlled experimental factor rather than a single default backbone. VGG-16 and ResNet-50 represent canonical convolutional architectures, with VGG-16 providing a plain deep CNN baseline and ResNet-50 representing residual CNNs. ViT-B/16 introduces a transformer-based visual encoder that operates over image patches rather than convolutional feature maps. CLIP ViT-B/32 and SigLIP add vision towers trained through paired image--text supervision, which is especially relevant for misinformation detection because the visual evidence must be interpreted relative to a textual claim. Holding the downstream pipeline fixed while varying these encoders allows us to test whether performance is driven mainly by convolutional versus transformer architecture, by unimodal versus multimodal pretraining, or by the specific visual representation learned by each backbone.

\subsubsection{Text Encoders}
We instantiate $\phi_t$ using five pre-trained language backbones: BERT-base~\cite{devlin2019bert}, DistilBERT~\cite{sanh2019distilbert}, SBERT~\cite{reimers2019sbert}, CLIP-Text~\cite{radford2021clip}, and SigLIP-Text~\cite{zhai2023siglip}. Together, these models cover standard bidirectional transformers, distilled encoders, sentence-oriented encoders, and text towers trained jointly with images.

This set is included to make the text encoder a separate controlled axis of the study. BERT-base provides a standard bidirectional Transformer baseline for contextual token-level language representations, while DistilBERT tests whether a smaller distilled encoder preserves enough semantic signal for the same downstream task. SBERT represents sentence-level representation learning, which is useful because misinformation examples are classified from complete claims or captions rather than isolated tokens. CLIP-Text and SigLIP-Text represent text towers trained under image--text alignment objectives, making them natural counterparts to multimodal vision encoders. Comparing these choices allows us to distinguish gains from general language understanding, sentence-level semantic embedding quality, and multimodal alignment.
\begin{figure}[t]
\centering
  \includegraphics[width=0.99\linewidth,height=0.49\linewidth]{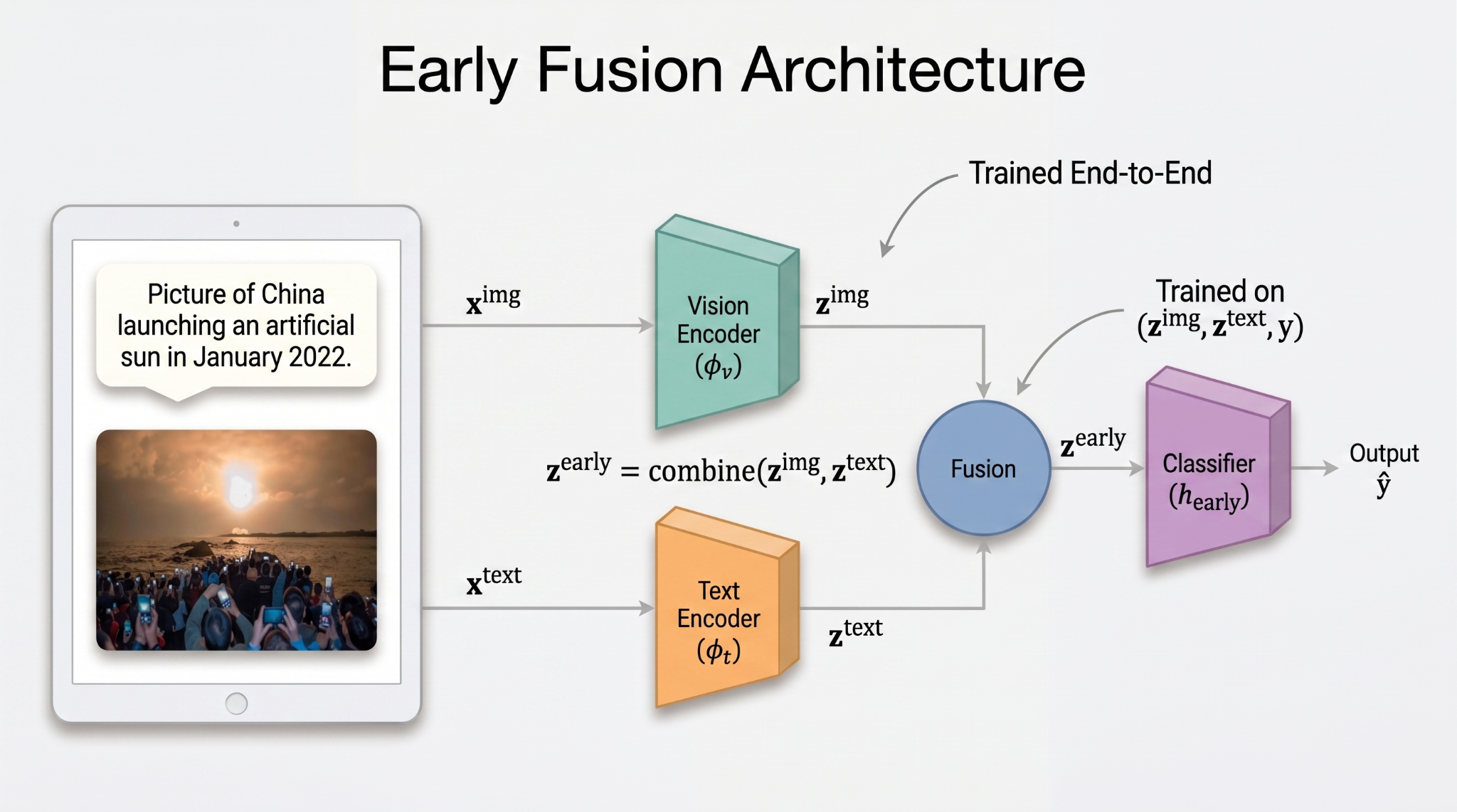}
  \caption{Figure showcasing the early fusion architecture used as part of our experiments}
  \label{fig:earlyfusionarch}
\end{figure}

\begin{figure*}[t]
\centering
  \includegraphics[width=0.99\linewidth,keepaspectratio]{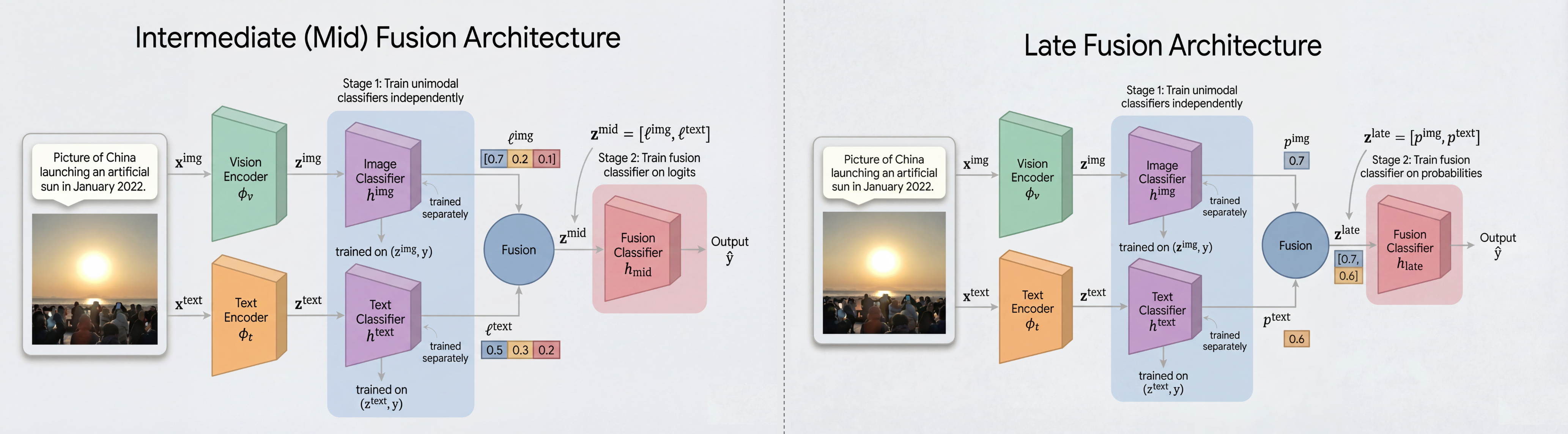}
   \caption{Figure showcasing the mid and late fusion architecture used as part of our experiments}
  \label{fig:midlatefusionarch}
\end{figure*}
\subsubsection{Classifiers}
We use three lightweight heads on top of frozen multimodal representations: logistic regression~\cite{cox1958logistic}, random forest~\cite{breiman2001random}, and XGBoost~\cite{chen2016xgboost}. This set provides a clean comparison between a standardized linear classifier, a bagging-based ensemble, and a boosting-based ensemble. Including logistic regression is important in our setting because frozen pre-trained representations can often be evaluated effectively with simple linear probes~\cite{kornblith2019better}, making a strong linear baseline difficult to beat.

These classifiers are included because the goal is to evaluate the representations and fusion strategies rather than to hide their effects behind a large trainable prediction head. Logistic regression provides the lowest-capacity linear probe: if it performs well, the frozen multimodal representation already exposes task-relevant information in a nearly linearly separable form. Random forest adds non-linear decision boundaries through bagged decision trees, while XGBoost provides a stronger boosting-based alternative that can capture feature interactions more aggressively. Using these three lightweight heads therefore gives an interpretable range of classifier capacity while keeping the main comparison focused on the encoders, fusion level, and modality dependence.
\subsubsection{Fusion Architectures}
Early fusion performs \textbf{feature-level} fusion by concatenating the image and text embeddings:
\[
\begin{aligned}
z^{\text{early}} &= [z^{\text{img}};z^{\text{text}}]\in\mathbb{R}^{d_v+d_t},\\
\hat{y} &= h_{\text{early}}(z^{\text{early}}).
\end{aligned}
\]
This is a single-stage scheme with no intermediate unimodal decision layer.

Mid fusion is a \textbf{score-level} two-stage scheme. We first train unimodal classifiers $h^{\text{img}}$ and $h^{\text{text}}$ and extract unimodal score representations $\ell^{\text{img}}, \ell^{\text{text}} \in \mathbb{R}^{C}$. These are concatenated and passed to a final fusion classifier:
\[
z^{\text{mid}}=[\ell^{\text{img}};\ell^{\text{text}}]\in\mathbb{R}^{2C},
\qquad
\hat{y}=h_{\text{mid}}(z^{\text{mid}}).
\]

Late fusion is a \textbf{decision-level} two-stage scheme. We first obtain unimodal probabilities $p^{\text{img}},p^{\text{text}}\in[0,1]^C$, concatenate them, and train a final classifier on the resulting vector:
\[
z^{\text{late}}=[p^{\text{img}};p^{\text{text}}]\in\mathbb{R}^{2C},
\qquad
\hat{y}=h_{\text{late}}(z^{\text{late}}).
\]
\subsection{Benchmark Setup}
We evaluate on three large-scale multimodal misinformation benchmarks: MMFakeBench, MiRAGeNews, and VLDBench~\cite{mmfakebench,miragenews,vldbench}. MMFakeBench contains 11,000 image–text pairs, MiRAGeNews contains 15,000 pairs across its training, validation, and test partitions, and VLDBench contains 31,339 article–image pairs. Together, they cover mixed-source content, AI-generated multimodal news, and vision–language disinformation.

For consistent supervised evaluation, we construct 80/20 train–test partitions for all three datasets, resulting in 8,800/2,200 examples for MMFakeBench, 12,000/3,000 for MiRAGeNews, and 25,071/6,268 for VLDBench. For MiRAGeNews, we remove 841 records with missing images—673 from the training partition and 168 from the test partition—leaving 14,159 instances (11,327/2,832). The final study comprises 56,498 examples: 11,000 from MMFakeBench, 14,159 from MiRAGeNews, and 31,339 from VLDBench. For cross-dataset transfer, we train on the constructed source training partition and evaluate on the target test partition.
\subsection{Implementation Details}

All encoders are frozen, and only the downstream classifier heads are trained. Logistic regression uses standardized inputs, the \texttt{lbfgs} solver, and \texttt{max\_iter}=2000; random forest uses 300 trees; and XGBoost uses 300 trees, a 0.05 learning rate, maximum depth 6, row and column subsampling of 0.8, histogram-based tree construction, and a binary logistic objective. Feature extraction uses batch sizes of 16 and 32 for vision and text, respectively, with a maximum text length of 64. All experiments run on an NVIDIA RTX A4000 GPU with 16\,GB VRAM.

Mid and late fusion use leakage-free stacking. For each outer training split, the unimodal branches generate out-of-fold predictions or features using shuffled 5-fold \texttt{StratifiedKFold} with a fixed random seed. The fusion classifier is trained only on these outputs, ensuring that no example both trains a unimodal model and contributes its in-sample prediction to the fusion model. The unimodal branches are then retrained on the full outer training split to generate outputs for the held-out test split, which is never used for training, calibration, or model selection. Cross-dataset transfer follows the same protocol: out-of-fold features and the fusion model are learned only from the source training set, and the frozen source-trained pipeline is evaluated directly on the target test set.

\noindent\textbf{Statistical analysis.}
For all four research questions, we report Macro-Precision, AUROC, and Accuracy with 95\% confidence intervals. Macro-Precision is the primary metric because false-positive credibility flags can be costly, while AUROC and Accuracy assess ranking quality and operating-point behavior. Aggregate comparisons over matched configurations use bootstrap confidence intervals. Paired comparisons use Wilcoxon signed-rank tests, with Holm correction for multiple pairwise tests; multi-level component comparisons use Friedman tests followed by Holm-corrected Wilcoxon post-hoc tests. We report Wilson score intervals for proportions, such as the frequency with which early fusion wins, and use exact McNemar tests for selected instance-level comparisons.

\section{Results and Discussion}
We organize the results by the four research questions. Unless otherwise noted, all table entries are mean percentages over matched configurations, and comparative claims such as ``best,'' ``strongest,'' or ``default'' refer to Macro-Precision, with AUROC and Accuracy reported as secondary diagnostics.

\subsection{RQ1: Generalized Performance Trends}

Based on Table~\ref{tab:fusion_summary_wide}, one in-domain trend is exceptionally stable: \textbf{early fusion is the only strategy that is best on average across all three datasets and metrics}. Over 225 matched in-domain configurations, it achieves 76.02\% Macro-Precision (95\% CI [74.58, 77.50]), 82.92\% AUROC ([81.46, 84.39]), and 76.87\% Accuracy ([75.51, 78.26]), compared with 74.23 / 81.64 / 76.04 for mid fusion and 74.09 / 81.37 / 75.91 for late fusion. Its Macro-Precision advantage is significant over both mid and late fusion ($p=1.00\times10^{-8}$ and $5.45\times10^{-11}$), and it wins 136/225 matched blocks, or 60.4\% with Wilson interval [53.9, 66.6].

Classifier rankings support exclusion rather than a single universal winner. \textbf{Random forest is the weakest head across all three datasets and metrics}, while logistic regression and XGBoost form the competitive pair in Table~\ref{tab:classifier_summary}. Logistic regression is nominally best in Macro-Precision at 76.42\%, but is tied with XGBoost at 76.34\% ($p=0.160$). XGBoost achieves the highest AUROC at 84.07\%, significantly above logistic regression at 83.33\% ($p=0.0022$), whereas logistic regression has the highest Accuracy at 78.08\%, with no significant difference from XGBoost's 77.93\% ($p=0.175$). XGBoost leads on MMFakeBench and VLDBench, while logistic regression leads on MiRAGeNews. Thus, random forest should be avoided, and the final choice should be between logistic regression and XGBoost depending on whether Macro-Precision or AUROC is prioritized.

\begin{table*}[t]
\centering
\small
\setlength{\tabcolsep}{7pt}
\begin{tabular}{llccc}
\toprule
Classifier & Metric & MMFakeBench & MiRAGeNews & VLDBench \\
\midrule
\multirow{3}{*}{Logistic Regression} & Prec & 73.93 ($\pm$ 1.07) & 91.65 ($\pm$ 0.95) & 63.68 ($\pm$ 0.48) \\
 & AUROC & 83.55 ($\pm$ 1.18) & 96.80 ($\pm$ 0.54) & 69.63 ($\pm$ 0.67) \\
 & Acc & 77.87 ($\pm$ 0.87) & 91.65 ($\pm$ 0.95) & 64.73 ($\pm$ 0.45) \\
\midrule
\multirow{3}{*}{XGBoost} & Prec & 74.51 ($\pm$ 0.74) & 90.00 ($\pm$ 0.83) & 64.52 ($\pm$ 0.32) \\
 & AUROC & 85.45 ($\pm$ 0.82) & 96.18 ($\pm$ 0.48) & 70.60 ($\pm$ 0.42) \\
 & Acc & 78.29 ($\pm$ 0.59) & 89.99 ($\pm$ 0.83) & 65.53 ($\pm$ 0.30) \\
\midrule
\multirow{3}{*}{Random Forest} & Prec & 71.06 ($\pm$ 1.31) & 83.34 ($\pm$ 1.29) & 60.32 ($\pm$ 0.68) \\
 & AUROC & 79.79 ($\pm$ 1.27) & 90.94 ($\pm$ 1.02) & 64.86 ($\pm$ 0.92) \\
 & Acc & 73.74 ($\pm$ 0.65) & 83.32 ($\pm$ 1.29) & 61.37 ($\pm$ 0.67) \\
\bottomrule
\end{tabular}
\caption{Classifier-wise in-domain averages (\%) over all encoder pairs and fusion strategies. Values are mean $\pm$ 95\% bootstrap CI half-width across matched configurations.}
\label{tab:classifier_summary}
\end{table*}

\begin{table*}[t]
\centering
\small
\setlength{\tabcolsep}{7pt}
\begin{tabular}{llccc}
\toprule
Fusion & Metric & MMFakeBench & MiRAGeNews & VLDBench \\
\midrule
\multirow{3}{*}{Early} & Prec & 75.20 ($\pm$ 1.07) & 88.80 ($\pm$ 1.15) & 64.06 ($\pm$ 0.41) \\
 & AUROC & 83.73 ($\pm$ 1.20) & 95.10 ($\pm$ 0.76) & 69.93 ($\pm$ 0.59) \\
 & Acc & 76.79 ($\pm$ 0.85) & 88.77 ($\pm$ 1.16) & 65.06 ($\pm$ 0.40) \\
\midrule
\multirow{3}{*}{Mid} & Prec & 72.22 ($\pm$ 1.07) & 88.20 ($\pm$ 1.41) & 62.27 ($\pm$ 0.71) \\
 & AUROC & 82.78 ($\pm$ 1.26) & 94.51 ($\pm$ 1.02) & 67.63 ($\pm$ 0.98) \\
 & Acc & 76.62 ($\pm$ 0.86) & 88.19 ($\pm$ 1.41) & 63.31 ($\pm$ 0.70) \\
\midrule
\multirow{3}{*}{Late} & Prec & 72.09 ($\pm$ 1.07) & 88.00 ($\pm$ 1.39) & 62.20 ($\pm$ 0.71) \\
 & AUROC & 82.28 ($\pm$ 1.22) & 94.31 ($\pm$ 0.99) & 67.52 ($\pm$ 0.96) \\
 & Acc & 76.48 ($\pm$ 0.85) & 87.99 ($\pm$ 1.38) & 63.25 ($\pm$ 0.69) \\
\bottomrule
\end{tabular}
\caption{Fusion-wise in-domain averages (\%) over all encoder pairs and classifier heads. Values are mean $\pm$ 95\% bootstrap CI half-width across matched configurations.}
\label{tab:fusion_summary_wide}
\end{table*}
The single-dataset evidence is strongest on VLDBench and MMFakeBench, where early fusion improves Macro-Precision by roughly two to three points over the best alternative, while preserving the same ordering on MiRAGeNews. This benchmark-consistent ranking shows that the result is not merely a small average effect across runs. When the evaluation target is unknown, early fusion is therefore the safest in-domain default for Macro-Precision, with AUROC and Accuracy providing consistent support.

\subsection{RQ2: Modality Complementarity}
The modality-drop analysis still supports complementarity, but the dominant pattern is not symmetric synergy; it is \textbf{image-heavy reliance}. Table~\ref{tab:modality_detailed} shows the full multimodal result together with both single-modality drop tests for every dataset--fusion--classifier block. Across all 675 matched comparisons, removing text reduces Macro-Precision by 4.83 points (95\% CI [4.18, 5.53]), AUROC by 3.93 ([3.65, 4.23]), and Accuracy by 5.78 ([5.28, 6.31]), whereas removing the image stream reduces them by 13.42 [12.20, 14.69], 12.08 [11.48, 12.66], and 16.93 [16.10, 17.79] points, respectively. Full multimodal models significantly outperform both text-drop and image-drop variants overall on all three metrics (all $p<10^{-78}$), and the difference between image-drop and text-drop degradation is also highly significant (Macro-Precision: $p=8.99\times10^{-46}$; AUROC: $p=2.26\times10^{-66}$; Accuracy: $p=1.31\times10^{-68}$).The asymmetry is strongly dataset dependent, but the direction is now uniform across all three benchmarks. MiRAGeNews is by far the most image-dependent benchmark: removing images causes mean losses of 21.68 points in Macro-Precision, 17.05 in AUROC, and 26.62 in Accuracy, far larger than the corresponding text-removal losses of 6.48, 3.27, and 8.52 points. MMFakeBench is the most balanced benchmark, although image removal is still worse on average (9.62 / 10.60 / 10.01 versus 6.61 / 6.42 / 5.33). VLDBench shows the smallest text losses and remains image-dominated as well (8.96 / 8.60 / 14.17 versus 1.39 / 2.11 / 3.49). Over all configurations, image removal hurts more than text removal in 76.3\% [72.9, 79.4] of Macro-Precision comparisons, 80.0\% [76.8, 82.8] of AUROC comparisons, and 82.4\% [79.3, 85.1] of Accuracy comparisons. The practical conclusion is therefore not merely that multimodal models help, but that their gains are real, asymmetric, and usually driven by harder-to-replace visual evidence.

\begin{table*}[t]
\centering
\small
\setlength{\tabcolsep}{3.0pt}
\begin{tabular}{@{}lllccccc@{}}
\toprule
Dataset & Fusion & Classifier & Full & Text drop & $\Delta_{\text{text}}$ & Image drop & $\Delta_{\text{img}}$ \\
\midrule
\multirow{9}{*}{MMFakeBench} & \multirow{3}{*}{Early} & Logistic Reg. & 72.2/82.0/76.6 & 64.4/75.2/68.2 & 7.8/6.8/8.5 & 62.8/71.0/65.7 & 9.4/11.0/11.0 \\
 &  & XGBoost & 76.2/86.2/79.0 & 71.0/79.2/72.5 & 5.2/7.1/6.5 & 63.4/72.9/65.7 & 12.9/13.3/13.3 \\
 &  & Random Forest & 77.2/83.0/74.7 & 74.2/76.8/72.8 & 3.0/6.1/1.9 & 58.4/72.3/71.0 & 18.8/10.7/3.8 \\
\cmidrule(lr){2-8}
 & \multirow{3}{*}{Mid} & Logistic Reg. & 74.9/84.9/78.6 & 66.4/78.0/72.3 & 8.4/6.9/6.3 & 69.2/77.7/73.2 & 5.7/7.2/5.4 \\
 &  & XGBoost & 73.8/85.2/78.0 & 69.5/81.0/73.6 & 4.3/4.2/4.4 & 67.2/76.2/61.9 & 6.6/9.0/16.2 \\
 &  & Random Forest & 68.0/78.2/73.2 & 62.3/70.4/68.5 & 5.7/7.8/4.8 & 58.0/65.1/64.2 & 10.0/13.1/9.0 \\
\cmidrule(lr){2-8}
 & \multirow{3}{*}{Late} & Logistic Reg. & 74.7/83.7/78.4 & 59.2/78.0/72.4 & 15.5/5.7/6.0 & 68.4/77.7/72.9 & 6.3/6.0/5.5 \\
 &  & XGBoost & 73.5/84.9/77.8 & 69.5/79.6/72.9 & 4.0/5.3/4.9 & 66.4/73.0/60.7 & 7.1/12.0/17.1 \\
 &  & Random Forest & 68.0/78.2/73.2 & 62.4/70.4/68.5 & 5.6/7.8/4.7 & 58.1/65.1/64.3 & 9.8/13.1/8.9 \\
\midrule
\multirow{9}{*}{MiRAGeNews} & \multirow{3}{*}{Early} & Logistic Reg. & 91.6/96.9/91.6 & 86.0/93.8/83.4 & 5.6/3.1/8.3 & 77.1/82.7/61.9 & 14.6/14.2/29.7 \\
 &  & XGBoost & 89.8/96.1/89.7 & 84.5/92.4/81.9 & 5.3/3.7/7.9 & 67.4/79.1/64.5 & 22.3/17.0/25.2 \\
 &  & Random Forest & 85.0/92.3/84.9 & 79.9/87.8/73.4 & 5.1/4.5/11.5 & 67.4/76.2/62.4 & 17.6/16.1/22.5 \\
\cmidrule(lr){2-8}
 & \multirow{3}{*}{Mid} & Logistic Reg. & 91.9/97.0/91.9 & 82.9/93.5/75.0 & 9.0/3.5/16.8 & 78.8/87.3/64.6 & 13.1/9.7/27.3 \\
 &  & XGBoost & 90.2/96.3/90.2 & 86.5/93.9/84.8 & 3.7/2.4/5.4 & 66.6/80.2/65.0 & 23.6/16.1/25.1 \\
 &  & Random Forest & 82.5/90.3/82.5 & 79.2/86.9/78.1 & 3.4/3.4/4.5 & 64.1/64.0/55.2 & 18.4/26.3/27.3 \\
\cmidrule(lr){2-8}
 & \multirow{3}{*}{Late} & Logistic Reg. & 91.4/96.5/91.4 & 72.1/93.5/78.6 & 19.4/3.0/12.8 & 47.9/87.3/60.7 & 43.6/9.2/30.8 \\
 &  & XGBoost & 90.0/96.2/90.0 & 86.5/93.7/85.0 & 3.6/2.5/5.1 & 66.5/77.7/65.6 & 23.6/18.5/24.4 \\
 &  & Random Forest & 82.5/90.3/82.5 & 79.1/86.9/78.0 & 3.4/3.4/4.5 & 64.1/63.9/55.3 & 18.4/26.3/27.2 \\
\midrule
\multirow{9}{*}{VLDBench} & \multirow{3}{*}{Early} & Logistic Reg. & 62.9/68.4/63.9 & 62.8/67.3/58.6 & 0.1/1.1/5.4 & 55.8/58.5/48.7 & 7.1/9.9/15.3 \\
 &  & XGBoost & 65.0/71.3/66.0 & 63.7/69.3/63.3 & 1.4/2.0/2.7 & 59.8/59.4/46.3 & 5.2/11.9/19.7 \\
 &  & Random Forest & 64.2/70.1/65.2 & 62.6/67.7/62.7 & 1.6/2.4/2.5 & 45.3/59.2/46.6 & 18.9/10.9/18.6 \\
\cmidrule(lr){2-8}
 & \multirow{3}{*}{Mid} & Logistic Reg. & 64.1/70.2/65.1 & 63.1/68.3/59.6 & 1.0/1.9/5.5 & 51.1/64.5/52.0 & 13.0/5.7/13.1 \\
 &  & XGBoost & 64.4/70.4/65.4 & 64.1/69.6/63.9 & 0.3/0.9/1.5 & 59.4/61.7/52.3 & 4.9/8.8/13.1 \\
 &  & Random Forest & 58.4/62.2/59.4 & 56.3/59.3/56.8 & 2.1/2.9/2.7 & 54.9/56.2/49.8 & 3.5/6.1/9.6 \\
\cmidrule(lr){2-8}
 & \multirow{3}{*}{Late} & Logistic Reg. & 64.1/70.3/65.1 & 61.6/68.3/60.1 & 2.5/2.0/5.0 & 46.8/64.5/51.4 & 17.3/5.8/13.7 \\
 &  & XGBoost & 64.2/70.0/65.2 & 62.7/67.4/61.7 & 1.5/2.7/3.4 & 57.0/57.9/50.3 & 7.1/12.2/14.8 \\
 &  & Random Forest & 58.3/62.2/59.4 & 56.3/59.3/56.8 & 2.0/3.0/2.6 & 54.8/56.1/49.8 & 3.5/6.1/9.7 \\
\bottomrule
\end{tabular}
\caption{Modality-drop study. Each cell reports Macro-Precision/AUROC/Accuracy (\%). ``Full'' is the complete multimodal model, and $\Delta_{\text{text}}$ and $\Delta_{\text{img}}$ are performance drops in percentage points when one modality is removed at test time. Detailed entries are rounded to one decimal for readability.}
\label{tab:modality_detailed}
\end{table*}

\subsection{RQ3: Component Impact Analysis}

The component-level picture remains stable and practically actionable under the revised metric protocol. All four component families exhibit significant overall effects under Friedman testing on Macro-Precision, AUROC, and Accuracy, but their magnitudes differ substantially. Vision encoder choice is dominant on all three metrics (all $p<8\times10^{-94}$), with average per-dataset spreads of 7.46 points on Macro-Precision, 7.34 on AUROC, and 6.95 on Accuracy. Classifier choice is the next largest source of stable variation (5.32 / 5.75 / 5.68 points), whereas text encoder choice (1.99 / 2.33 / 1.70) and fusion strategy (1.93 / 1.55 / 0.97) are materially smaller. Put differently, once a reasonable text encoder is in place, most of the robust variance comes from the image tower and the downstream head.

Two trends are especially stable. First, \textbf{SigLIP is the strongest vision encoder on all three reported metrics}, reaching 79.19\% Macro-Precision, 86.01\% AUROC, and 80.37\% Accuracy overall, with CLIP ViT-B/32 ranking second at 77.43 / 84.93 / 78.67. Post-hoc testing confirms that SigLIP significantly exceeds CLIP ViT-B/32 after Holm correction on all three metrics ($p_{\text{Holm}}=7.40\times10^{-15}$, $2.53\times10^{-20}$, and $5.67\times10^{-15}$). Second, \textbf{text encoders matter less and separate much less cleanly}. DistilBERT has the strongest overall in-domain text means at 75.17 / 82.57 / 76.67, but the top four text encoders are all within 0.32 Macro-Precision points of one another, and DistilBERT does not significantly exceed BERT-base or CLIP-Text after Holm correction. This is exactly the pattern one would expect from a component family that matters, but matters less than the vision side.

\begin{figure*}[t]

\centering

\includegraphics[width=0.99\textwidth,keepaspectratio]{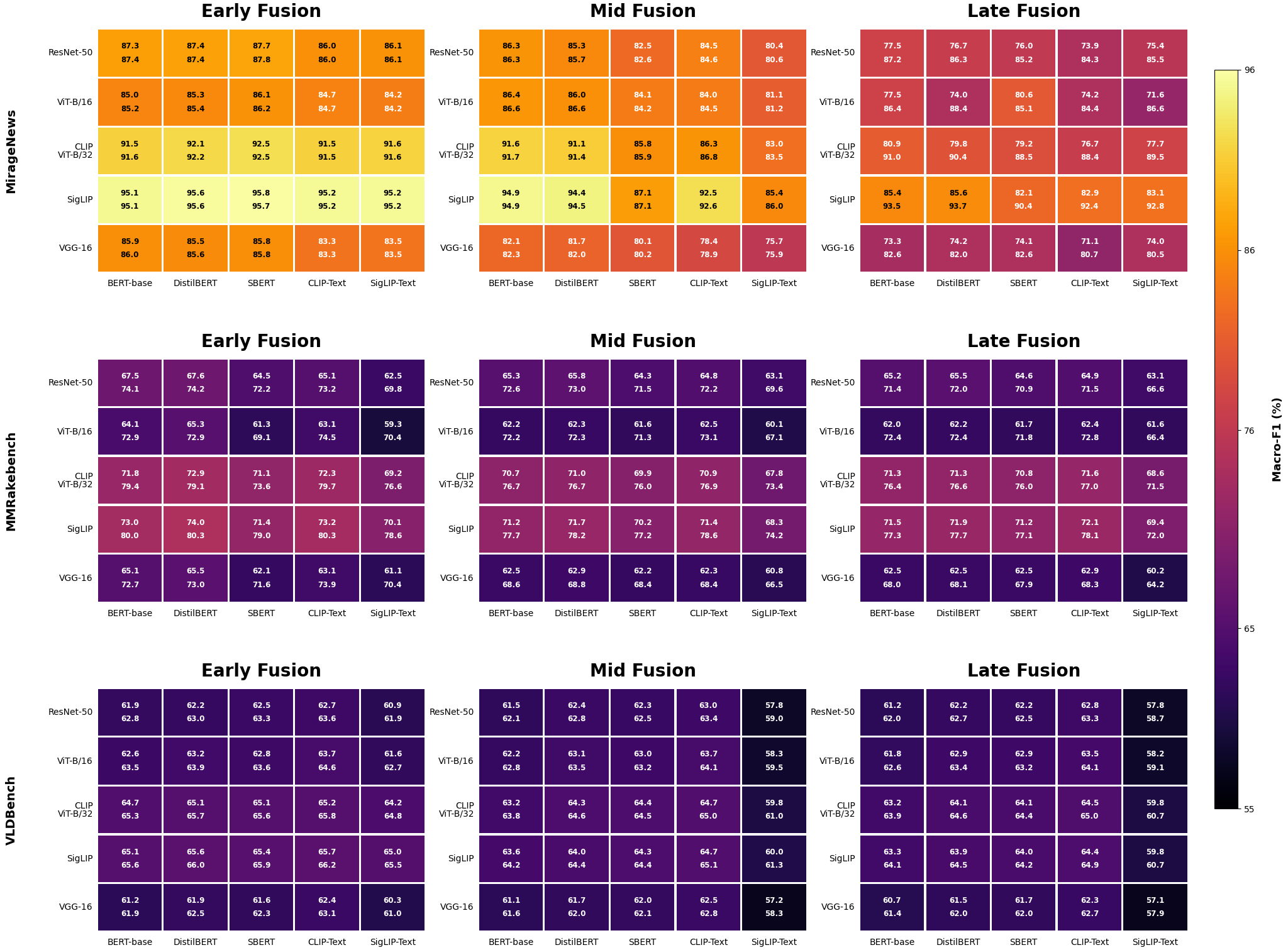}

\caption{Main encoder-pair heatmap across the three datasets. Each cell reports mean Macro-Precision / Accuracy (\%) for a fixed vision--text encoder pair, averaged over the three classifier heads under the given fusion strategy; color encodes AUROC.}

\label{fig:main_heatmap}

\end{figure*}

Classifier conclusions mirror RQ1. Logistic regression and XGBoost again form the leading pair: logistic regression has the highest overall Macro-Precision and Accuracy means (76.42 and 78.08), but both are statistically tied with XGBoost ($p=0.160$ and $0.175$), whereas XGBoost has the strongest AUROC at 84.07\%, significantly above logistic regression ($p=0.0022$). Random forest is substantially behind on every metric. Figure~\ref{fig:main_heatmap} reinforces the same conclusion visually: variation across vision rows is larger and more systematic than variation across most text columns. Taken together, the component analysis suggests a clear priority order for future system design: get the vision encoder right first, choose between logistic regression and XGBoost as the serious downstream heads, then refine the fusion strategy, and only then expect modest gains from swapping among strong text encoders such as DistilBERT, BERT-base, CLIP-Text, and SBERT.

\subsection{RQ4: Benchmark-Shift Generalization}

Benchmark shift is the most challenging setting. As Table~\ref{tab:transfer_direction} shows, all six source--target transfers degrade substantially relative to their target in-domain baselines. Macro-Precision gaps range from 8.33 points for MMFakeBench$\rightarrow$VLDBench to 36.30 for VLDBench$\rightarrow$MiRAGeNews; the corresponding AUROC and Accuracy gaps range from 11.76 to 41.57 and from 6.45 to 36.93 points, respectively. The highest transfer Macro-Precision is achieved by MMFakeBench$\rightarrow$MiRAGeNews at 63.25\% ($\pm$0.59), followed by MiRAGeNews$\rightarrow$MMFakeBench at 62.56\% ($\pm$0.53), whereas VLDBench$\rightarrow$MiRAGeNews exhibits the largest overall collapse. Every transfer direction is significantly worse than its target in-domain baseline on all three metrics (paired Wilcoxon tests, all $p\leq1.15\times10^{-38}$).

\begin{table*}[t]
\centering
\setlength{\tabcolsep}{7pt}
\small
\begin{tabular}{lccc}
\toprule
Train $\rightarrow$ Test & Macro-Precision & AUROC & Accuracy \\
\midrule
MMFakeBench $\rightarrow$ MiRAGeNews & 63.25 ($\pm$ 0.59) & 67.70 ($\pm$ 0.70) & 57.84 ($\pm$ 0.57) \\
MMFakeBench $\rightarrow$ VLDBench & 54.51 ($\pm$ 0.33) & 56.60 ($\pm$ 0.25) & 57.42 ($\pm$ 0.14) \\
MiRAGeNews $\rightarrow$ MMFakeBench & 62.56 ($\pm$ 0.53) & 69.26 ($\pm$ 0.70) & 55.44 ($\pm$ 0.70) \\
MiRAGeNews $\rightarrow$ VLDBench & 49.25 ($\pm$ 0.15) & 48.88 ($\pm$ 0.22) & 44.85 ($\pm$ 0.19) \\
VLDBench $\rightarrow$ MMFakeBench & 52.64 ($\pm$ 0.26) & 54.69 ($\pm$ 0.46) & 48.68 ($\pm$ 0.39) \\
VLDBench $\rightarrow$ MiRAGeNews & 52.03 ($\pm$ 0.35) & 53.07 ($\pm$ 0.52) & 51.39 ($\pm$ 0.27) \\
\bottomrule
\end{tabular}
\caption{Cross-dataset transfer by source--target direction. Values are mean $\pm$ 95\% bootstrap CI half-width (\%).}
\label{tab:transfer_direction}
\end{table*}

\begin{table*}[t]
\centering
\setlength{\tabcolsep}{6pt}
\small
\begin{tabular}{llccc}
\toprule
\multicolumn{5}{c}{(a) Classifier-wise transfer averages by target dataset} \\
\midrule
Classifier & Metric & MMFakeBench & MiRAGeNews & VLDBench \\
\midrule
\multirow{3}{*}{Logistic Regression} & Prec & 58.22 ($\pm$ 0.97) & 58.77 ($\pm$ 1.02) & 51.64 ($\pm$ 0.52) \\
 & AUROC & 63.10 ($\pm$ 1.46) & 62.02 ($\pm$ 1.32) & 52.80 ($\pm$ 0.59) \\
 & Acc & 52.38 ($\pm$ 0.85) & 55.21 ($\pm$ 0.73) & 51.00 ($\pm$ 1.05) \\
\midrule
\multirow{3}{*}{XGBoost} & Prec & 58.22 ($\pm$ 1.01) & 58.45 ($\pm$ 1.12) & 52.26 ($\pm$ 0.55) \\
 & AUROC & 62.62 ($\pm$ 1.39) & 61.89 ($\pm$ 1.53) & 52.82 ($\pm$ 0.74) \\
 & Acc & 51.89 ($\pm$ 1.02) & 55.99 ($\pm$ 0.84) & 51.24 ($\pm$ 1.07) \\
\midrule
\multirow{3}{*}{Random Forest} & Prec & 56.36 ($\pm$ 0.87) & 55.70 ($\pm$ 1.05) & 51.73 ($\pm$ 0.51) \\
 & AUROC & 60.19 ($\pm$ 1.28) & 57.24 ($\pm$ 1.22) & 52.60 ($\pm$ 0.69) \\
 & Acc & 51.92 ($\pm$ 0.79) & 52.64 ($\pm$ 0.58) & 51.17 ($\pm$ 0.93) \\
\bottomrule
\end{tabular}

\vspace{6pt}

\begin{tabular}{llccc}
\toprule
\multicolumn{5}{c}{(b) Fusion-wise transfer averages by target dataset} \\
\midrule
Fusion & Metric & MMFakeBench & MiRAGeNews & VLDBench \\
\midrule
\multirow{3}{*}{Early} & Prec & 57.69 ($\pm$ 0.92) & 58.74 ($\pm$ 1.21) & 52.04 ($\pm$ 0.57) \\
 & AUROC & 62.28 ($\pm$ 1.33) & 60.96 ($\pm$ 1.49) & 53.09 ($\pm$ 0.70) \\
 & Acc & 52.55 ($\pm$ 0.92) & 53.92 ($\pm$ 0.73) & 51.50 ($\pm$ 0.99) \\
\midrule
\multirow{3}{*}{Mid} & Prec & 57.59 ($\pm$ 0.96) & 57.11 ($\pm$ 1.00) & 51.82 ($\pm$ 0.49) \\
 & AUROC & 61.90 ($\pm$ 1.38) & 60.19 ($\pm$ 1.35) & 52.58 ($\pm$ 0.66) \\
 & Acc & 51.74 ($\pm$ 0.86) & 54.98 ($\pm$ 0.77) & 50.91 ($\pm$ 1.02) \\
\midrule
\multirow{3}{*}{Late} & Prec & 57.51 ($\pm$ 0.94) & 57.07 ($\pm$ 0.99) & 51.78 ($\pm$ 0.52) \\
 & AUROC & 61.74 ($\pm$ 1.35) & 60.01 ($\pm$ 1.33) & 52.55 ($\pm$ 0.68) \\
 & Acc & 51.90 ($\pm$ 0.84) & 54.95 ($\pm$ 0.77) & 50.99 ($\pm$ 1.02) \\
\bottomrule
\end{tabular}
\caption{Cross-dataset transfer averages by component family and target dataset. Values are mean $\pm$ 95\% bootstrap CI half-width (\%).}
\label{tab:transfer_breakdown}
\end{table*}
Although transfer rankings are flatter than in-domain rankings, early fusion remains the strongest precision-first default. Across 450 matched transfer blocks, it achieves 56.16\% Macro-Precision, significantly exceeding mid fusion (55.51\%, $p=1.73\times10^{-5}$) and late fusion (55.45\%, $p=1.47\times10^{-5}$). It also leads in AUROC at 58.78\%, while Accuracy is nearly tied across fusion strategies. Direction-specific reversals occur---MiRAGeNews$\rightarrow$MMFakeBench slightly favors mid fusion, and VLDBench$\rightarrow$MiRAGeNews favors late fusion---but do not alter the aggregate ranking.
Table~\ref{tab:transfer_breakdown} shows that other component rankings also flatten under transfer. XGBoost has the highest Macro-Precision at 56.31\% but is statistically tied with logistic regression at 56.21\% ($p=0.264$). SigLIP leads the vision encoders at 57.73\% but is tied with CLIP ViT-B/32 at 57.49\% ($p=0.819$), while BERT-base leads the less-stable text encoders at 56.00\%, only 0.19 points above SBERT.

\textbf{Overall, the safest precision-first transfer configuration uses early fusion, SigLIP or CLIP ViT-B/32, BERT-base or SBERT, and either XGBoost or logistic regression.} Source-specific effects may change the finer ranking, but strong visual representations and early fusion remain the most dependable choices. Cross-dataset evaluation should therefore be treated as a first-class requirement, as even strong in-domain models can lose more than 30 points under benchmark shift.
\subsection{SHAP-Based Interpretation}
We also use SHAP to test whether the modality conclusions are visible in feature attributions rather than only in modality-drop ablations. Across 675 in-domain SHAP analyses, image features account for a mean 61.57\% of total attribution mass (95\% bootstrap CI [60.52, 62.61]). Image attribution exceeds text attribution in 81.6\% of configurations, and the paired image-vs-text difference is highly significant ($p=7.24\times10^{-50}$). This pattern is not a fusion-specific artifact: the mean image share is essentially identical for early, mid, and late fusion (all about 61.6\%).

The dataset breakdown mirrors RQ2 almost exactly. MMFakeBench is the most balanced benchmark at 53.8\% mean image share, MiRAGeNews is the most visually dominated at 67.1\%, and VLDBench remains clearly image-leaning at 63.8\%. In the best-performing SHAP-analyzed configurations for each dataset, the top individual contributors are overwhelmingly SigLIP dimensions, which independently reinforces the RQ3 conclusion that visual encoder quality---and especially strong contrastive vision backbones---explains a large share of stable multimodal performance.
\subsection{Practical Recommendations}
The aggregate results point to two concrete default approaches for future multimodal misinformation studies. If the goal is strong in-domain performance under frozen representations, a robust Macro-Precision-oriented starting point is SigLIP on the visual side, with CLIP ViT-B/32 as the main alternative, DistilBERT on the text side, early fusion, and either logistic regression or XGBoost as the downstream head. This recommendation is supported by the in-domain averages: early fusion is the strongest fusion strategy overall on Macro-Precision (76.02\%), SigLIP is the strongest vision encoder (79.19\%), DistilBERT has the strongest text-encoder mean (75.17\%), and logistic regression and XGBoost form the leading classifier pair, with logistic regression nominally highest on Macro-Precision (76.42\%) and XGBoost essentially tied (76.34\%).

If the focus is stronger OOD performance, the default should be chosen for transfer stability rather than in-domain dominance. Our results suggest early fusion with a strong contrastive vision encoder, especially SigLIP or CLIP ViT-B/32, together with a strong generic text encoder such as BERT-base or SBERT, and either XGBoost or logistic regression as the final head. This recommendation follows the transfer-wide Macro-Precision ranking: early fusion is best overall (56.16\%), SigLIP is the strongest vision encoder (57.73\%) but statistically tied with CLIP ViT-B/32 (57.49\%), BERT-base is the strongest text encoder (56.00\%) with SBERT close behind (55.81\%), and XGBoost is nominally best among classifiers (56.31\%) but statistically tied with logistic regression (56.21\%). The main shift from the in-domain recommendation is therefore on the text side: DistilBERT is the strongest in-domain text choice, whereas under benchmark shift the more reliable text defaults become BERT-base and SBERT.

\subsection{Limitations and Future Work}
Our study intentionally focuses on frozen encoders and simple downstream heads. This choice improves comparability and isolates component effects, but it does not cover fully trainable cross-attention architectures or parameter-efficient adaptation methods.  Finally, statistical significance should not be confused with practical magnitude: because we aggregate many matched configurations, some small gaps are estimated precisely without necessarily justifying additional system complexity. Future work should therefore extend the benchmark scope and pair significance testing with stronger effect-size, calibration, and more in-depth analyses.

\section{Conclusion}
In this paper, we presented a large-scale empirical study of multimodal misinformation detection, covering 3,375 controlled experiments across three benchmarks, multiple vision and text encoders, fusion strategies, and classifiers. Our findings show that early fusion is relatively the most reliable in-domain strategy, visual evidence often contributes more strongly than text, and vision encoder and classifier choices have the largest impact on performance. We also find that cross-dataset generalization remains challenging, highlighting the need for stronger transfer evaluation. We hope that these findings and protocols will serve as a useful reference for the broader research community.

\bibliography{references}

@inproceedings{he2016resnet,
  author = {He, Kaiming and Zhang, Xiangyu and Ren, Shaoqing and Sun, Jian},
  title = {Deep Residual Learning for Image Recognition},
  booktitle = {Proceedings of the IEEE Conference on Computer Vision and Pattern Recognition},
  pages = {770--778},
  year = {2016},
  doi = {10.1109/CVPR.2016.90}
}

@inproceedings{dosovitskiy2021vit,
  author = {Dosovitskiy, Alexey and Beyer, Lucas and Kolesnikov, Alexander and Weissenborn, Dirk and Zhai, Xiaohua and Unterthiner, Thomas and Dehghani, Mostafa and Minderer, Matthias and Heigold, Georg and Gelly, Sylvain and Uszkoreit, Jakob and Houlsby, Neil},
  title = {An Image is Worth 16x16 Words: Transformers for Image Recognition at Scale},
  booktitle = {International Conference on Learning Representations},
  year = {2021}
}

@inproceedings{radford2021clip,
  author = {Radford, Alec and Kim, Jong Wook and Hallacy, Chris and Ramesh, Aditya and Goh, Gabriel and Agarwal, Sandhini and Sastry, Girish and Askell, Amanda and Mishkin, Pamela and Clark, Jack and Krueger, Gretchen and Sutskever, Ilya},
  title = {Learning Transferable Visual Models From Natural Language Supervision},
  booktitle = {Proceedings of the 38th International Conference on Machine Learning},
  series = {Proceedings of Machine Learning Research},
  volume = {139},
  pages = {8748--8763},
  publisher = {PMLR},
  year = {2021}
}

@inproceedings{zhai2023siglip,
  author = {Zhai, Xiaohua and Mustafa, Basil and Kolesnikov, Alexander and Beyer, Lucas},
  title = {Sigmoid Loss for Language Image Pre-Training},
  booktitle = {Proceedings of the IEEE/CVF International Conference on Computer Vision},
  pages = {11975--11986},
  year = {2023}
}

@inproceedings{simonyan2015vgg,
  author = {Simonyan, Karen and Zisserman, Andrew},
  title = {Very Deep Convolutional Networks for Large-Scale Image Recognition},
  booktitle = {International Conference on Learning Representations},
  year = {2015}
}

@inproceedings{devlin2019bert,
  author = {Devlin, Jacob and Chang, Ming-Wei and Lee, Kenton and Toutanova, Kristina},
  title = {{BERT}: Pre-training of Deep Bidirectional Transformers for Language Understanding},
  booktitle = {Proceedings of the 2019 Conference of the North American Chapter of the Association for Computational Linguistics: Human Language Technologies},
  pages = {4171--4186},
  year = {2019},
  doi = {10.18653/v1/N19-1423}
}

@misc{sanh2019distilbert,
  author = {Sanh, Victor and Debut, Lysandre and Chaumond, Julien and Wolf, Thomas},
  title = {{DistilBERT}, a Distilled Version of {BERT}: Smaller, Faster, Cheaper and Lighter},
  year = {2019},
  eprint = {1910.01108},
  archivePrefix = {arXiv},
  primaryClass = {cs.CL}
}

@inproceedings{reimers2019sbert,
  author = {Reimers, Nils and Gurevych, Iryna},
  title = {Sentence-{BERT}: Sentence Embeddings Using Siamese {BERT}-Networks},
  booktitle = {Proceedings of the 2019 Conference on Empirical Methods in Natural Language Processing and the 9th International Joint Conference on Natural Language Processing},
  pages = {3982--3992},
  year = {2019},
  doi = {10.18653/v1/D19-1410}
}

@article{cox1958logistic,
  author = {Cox, D. R.},
  title = {The Regression Analysis of Binary Sequences},
  journal = {Journal of the Royal Statistical Society: Series B (Methodological)},
  volume = {20},
  number = {2},
  pages = {215--232},
  year = {1958},
  doi = {10.1111/j.2517-6161.1958.tb00292.x}
}

@article{breiman2001random,
  author = {Breiman, Leo},
  title = {Random Forests},
  journal = {Machine Learning},
  volume = {45},
  pages = {5--32},
  year = {2001},
  doi = {10.1023/A:1010933404324}
}

@inproceedings{chen2016xgboost,
  author = {Chen, Tianqi and Guestrin, Carlos},
  title = {{XGBoost}: A Scalable Tree Boosting System},
  booktitle = {Proceedings of the 22nd ACM SIGKDD International Conference on Knowledge Discovery and Data Mining},
  pages = {785--794},
  year = {2016},
  doi = {10.1145/2939672.2939785}
}

@inproceedings{kornblith2019better,
  author = {Kornblith, Simon and Shlens, Jonathon and Le, Quoc V.},
  title = {Do Better {ImageNet} Models Transfer Better?},
  booktitle = {Proceedings of the IEEE/CVF Conference on Computer Vision and Pattern Recognition},
  pages = {2661--2671},
  year = {2019},
  doi = {10.1109/CVPR.2019.00277}
}

@article{abdali2024multimodal,
  author = {Abdali, Sara and Shaham, Sina and Krishnamachari, Bhaskar},
  title = {Multi-modal Misinformation Detection: Approaches, Challenges and Opportunities},
  journal = {ACM Computing Surveys},
  volume = {57},
  number = {3},
  pages = {76:1--76:29},
  year = {2024},
  doi = {10.1145/3697349}
}

@article{comito2023survey,
  author = {Comito, Carmela and Caroprese, Luciano and Zumpano, Ester},
  title = {Multimodal Fake News Detection on Social Media: A Survey of Deep Learning Techniques},
  journal = {Social Network Analysis and Mining},
  volume = {13},
  number = {1},
  pages = {101},
  year = {2023},
  doi = {10.1007/s13278-023-01104-w}
}

@inproceedings{jin2017attrnn,
  author = {Jin, Zhiwei and Cao, Juan and Guo, Han and Zhang, Yongdong and Luo, Jiebo},
  title = {Multimodal Fusion with Recurrent Neural Networks for Rumor Detection on Microblogs},
  booktitle = {Proceedings of the 25th ACM International Conference on Multimedia},
  pages = {795--816},
  year = {2017},
  doi = {10.1145/3123266.3123454}
}

@inproceedings{wang2018eann,
  author = {Wang, Yaqing and Ma, Fenglong and Jin, Zhiwei and Yuan, Ye and Xun, Guangxu and Jha, Kishlay and Su, Lu and Gao, Jing},
  title = {{EANN}: Event Adversarial Neural Networks for Multi-modal Fake News Detection},
  booktitle = {Proceedings of the 24th ACM SIGKDD International Conference on Knowledge Discovery and Data Mining},
  pages = {849--857},
  year = {2018},
  doi = {10.1145/3219819.3219903}
}

@inproceedings{singhal2019spotfake,
  author = {Singhal, Shivangi and Shah, Rajiv Ratn and Chakraborty, Tanmoy and Kumaraguru, Ponnurangam and Satoh, Shin'ichi},
  title = {{SpotFake}: A Multi-modal Framework for Fake News Detection},
  booktitle = {2019 IEEE Fifth International Conference on Multimedia Big Data (BigMM)},
  pages = {39--47},
  year = {2019},
  doi = {10.1109/BigMM.2019.00-44}
}

@inproceedings{singhal2020spotfakeplus,
  author = {Singhal, Shivangi and Kabra, Anubha and Sharma, Mohit and Shah, Rajiv Ratn and Chakraborty, Tanmoy and Kumaraguru, Ponnurangam},
  title = {{SpotFake+}: A Multimodal Framework for Fake News Detection via Transfer Learning},
  booktitle = {Proceedings of the AAAI Conference on Artificial Intelligence},
  volume = {34},
  number = {10},
  pages = {13915--13916},
  year = {2020},
  doi = {10.1609/aaai.v34i10.7230}
}

@inproceedings{khattar2019mvae,
  author = {Khattar, Dhruv and Goud, Jaipal Singh and Gupta, Manish and Varma, Vasudeva},
  title = {{MVAE}: Multimodal Variational Autoencoder for Fake News Detection},
  booktitle = {The World Wide Web Conference},
  pages = {2915--2921},
  year = {2019},
  doi = {10.1145/3308558.3313552}
}

@incollection{zhou2020safe,
  author = {Zhou, Xinyi and Wu, Jindi and Zafarani, Reza},
  title = {{SAFE}: Similarity-Aware Multi-modal Fake News Detection},
  booktitle = {Advances in Knowledge Discovery and Data Mining},
  series = {Lecture Notes in Computer Science},
  volume = {12085},
  pages = {354--367},
  publisher = {Springer},
  address = {Cham},
  year = {2020},
  doi = {10.1007/978-3-030-47436-2_27}
}

@article{xue2021mcnn,
  author = {Xue, Junxiao and Wang, Yabo and Tian, Yichen and Li, Yafei and Shi, Lei and Wei, Lin},
  title = {Detecting Fake News by Exploring the Consistency of Multimodal Data},
  journal = {Information Processing \& Management},
  volume = {58},
  number = {5},
  pages = {102610},
  year = {2021},
  doi = {10.1016/j.ipm.2021.102610}
}

@inproceedings{luo2021newsclippings,
  author = {Luo, Grace and Darrell, Trevor and Rohrbach, Anna},
  title = {{NewsCLIPpings}: Automatic Generation of Out-of-Context Multimodal Media},
  booktitle = {Proceedings of the 2021 Conference on Empirical Methods in Natural Language Processing},
  pages = {6801--6817},
  year = {2021},
  doi = {10.18653/v1/2021.emnlp-main.545}
}

@inproceedings{qi2024sniffer,
  author = {Qi, Peng and Yan, Zehong and Hsu, Wynne and Lee, Mong Li},
  title = {{SNIFFER}: Multimodal Large Language Model for Explainable Out-of-Context Misinformation Detection},
  booktitle = {Proceedings of the IEEE/CVF Conference on Computer Vision and Pattern Recognition},
  pages = {13052--13062},
  year = {2024},
  eprint = {2403.03170},
  archivePrefix = {arXiv}
}

@inproceedings{mmfakebench,
  author = {Liu, Xuannan and Li, Zekun and Li, Pei and Huang, Huaibo and Xia, Shuhan and Cui, Xing and Huang, Linzhi and Deng, Weihong and He, Zhaofeng},
  title = {{MMFakeBench}: A Mixed-Source Multimodal Misinformation Detection Benchmark for {LVLMs}},
  booktitle = {International Conference on Learning Representations},
  year = {2025}
}

@inproceedings{miragenews,
  author = {Huang, Runsheng and Dugan, Liam and Yang, Yue and Callison-Burch, Chris},
  title = {{MiRAGeNews}: Multimodal Realistic {AI}-Generated News Detection},
  booktitle = {Findings of the Association for Computational Linguistics: EMNLP 2024},
  pages = {16436--16448},
  year = {2024},
  address = {Miami, Florida, USA},
  publisher = {Association for Computational Linguistics},
  doi = {10.18653/v1/2024.findings-emnlp.959}
}

@article{vldbench,
  author = {Raza, Shaina and Vayani, Ashmal and Jain, Aditya and Narayanan, Aravind and Khazaie, Vahid Reza and Bashir, Syed Raza and Dolatabadi, Elham and Uddin, Gias and Emmanouilidis, Christos and Qureshi, Rizwan and Shah, Mubarak},
  title = {{VLDBench} Evaluating Multimodal Disinformation with Regulatory Alignment},
  journal = {Information Fusion},
  volume = {130},
  pages = {104092},
  year = {2026},
  doi = {10.1016/j.inffus.2025.104092}
}

@article{chen2023robustness,
  author = {Chen, Jinyin and Jia, Chengyu and Zheng, Haibin and Chen, Ruoxi and Fu, Chenbo},
  title = {Is Multi-modal Necessarily Better? Robustness Evaluation of Multi-modal Fake News Detection},
  journal = {IEEE Transactions on Network Science and Engineering},
  volume = {10},
  pages = {3144--3158},
  year = {2023},
  doi = {10.1109/TNSE.2023.3249290}
}

@inproceedings{kumar2022hateclipper,
  author = {Kumar, Gokul Karthik and Nandakumar, Karthik},
  title = {{Hate-CLIPper}: Multimodal Hateful Meme Classification Based on Cross-Modal Interaction of {CLIP} Features},
  booktitle = {Proceedings of the Second Workshop on {NLP} for Positive Impact ({NLP4PI})},
  pages = {171--183},
  address = {Abu Dhabi, United Arab Emirates (Hybrid)},
  publisher = {Association for Computational Linguistics},
  month = dec,
  year = {2022},
  doi = {10.18653/v1/2022.nlp4pi-1.20},
  url = {https://aclanthology.org/2022.nlp4pi-1.20/}
}

@inproceedings{shah2024memeclip,
  author = {Shah, Siddhant Bikram and Shiwakoti, Shuvam and Chaudhary, Maheep and Wang, Haohan},
  title = {{MemeCLIP}: Leveraging {CLIP} Representations for Multimodal Meme Classification},
  booktitle = {Proceedings of the 2024 Conference on Empirical Methods in Natural Language Processing},
  pages = {17320--17332},
  address = {Miami, Florida, USA},
  publisher = {Association for Computational Linguistics},
  month = nov,
  year = {2024},
  doi = {10.18653/v1/2024.emnlp-main.959},
  url = {https://aclanthology.org/2024.emnlp-main.959/}
}

@inproceedings{sharma2026memetag,
  author = {Sharma, Akshit and Patil, Prashant W.},
  title = {{MemeTAG}: Keyword-Driven Meme Classification through Tag Embedding Reconstruction},
  booktitle = {Proceedings of the IEEE/CVF Winter Conference on Applications of Computer Vision ({WACV})},
  pages = {7679--7688},
  month = mar,
  year = {2026},
  doi = {10.1109/WACV61042.2026.00741},
  url = {https://doi.org/10.1109/WACV61042.2026.00741}
}

@misc{sharma2026smallcues,
  author = {Sharma, Akshit and Patil, Prashant W.},
  title = {Small Cues, Big Consequences: Learning Pivotal Cues for Multimodal Meme Classification},
  year = {2026},
  eprint = {2609.26907},
  archivePrefix = {arXiv},
  primaryClass = {cs.MM},
  note = {Accepted to Findings of the Association for Computational Linguistics: {EMNLP} 2026},
  url = {https://arxiv.org/abs/2609.26907},
  doi = {10.48550/arXiv.2609.26907}
}

\end{document}